\documentclass[usletter, 10pt, conference]{article} 

\usepackage{genom}
\usepackage{bigfoot}
\DeclareNewFootnote[para]{default}
\usepackage{times}

\usepackage{common}
\usepackage{listings}
\usepackage{lstlang3}
\usepackage{amsmath, amssymb}
\usepackage{indentfirst}

\usepackage{inconsolata}
\usepackage{color}
\definecolor{bluekeywords}{rgb}{0.13,0.13,1}
\definecolor{greencomments}{rgb}{0,0.5,0}
\definecolor{redstrings}{rgb}{0.9,0,0}
\usepackage{latexsym}
\usepackage{units}

\usepackage{hyperref}

\usepackage{enumitem}

\usepackage{todonotes}
\newcommand{\comment}[1]{ }

\newcommand{\task}[1]{\textsf{#1}}
\newcommand{\port}[1]{\textbf{#1}}
\newcommand{\modu}[1]{\textsc{#1}}
\newcommand{\serv}[1]{\textit{#1}}

\newcommand{\codel}[1]{\texttt{#1}}

\newcommand{\tina}[0]{TINA\xspace}
\newcommand{\fcr}[0]{Fiacre\xspace}

\title{\LARGE \bf
\GenoMMM{}  Templates: from Middleware Independence to \\Formal Models Synthesis
 \thanks{This work was supported in part by the EU CPSE Labs project funded by the H2020 program under grant agreement No 644400.}
} 

\author{Mohammed Foughali \hspace{1cm} F\'{e}lix Ingrand \hspace{1cm} Anthony Mallet\\
LAAS-CNRS, Universit\'e de Toulouse, CNRS, Toulouse, France\\
{\tt\small \{firstname.lastname\}@laas.fr}}%

\date{}
\begin{document}
\sloppy

\renewcommand{\thelstlisting}{\arabic{lstlisting}}

\maketitle

\begin{abstract} 

  \GenoM{} is an approach to develop robotic software components, which can be controlled, and assembled to build complex applications. Its
  latest version \GenoMMM{}, provides a template mechanism which is versatile enough to deploy components for different middleware without
  any change in the specification and user code. But this same template mechanism also enables us to automatically synthesize formal models
  (for two Validation and Verification frameworks) of the final components. We illustrate our approach on a real deployed example of a drone
  flight controller for which we prove offline real-time properties, and an outdoor robot for which we synthesize a controller to perform
  runtime verification.

\end{abstract}

\section{Introduction}
\label{sec:intro}
There is a rising concern in advanced robotic and autonomous systems software development. Can we improve the dependability of such systems
by deploying formal validation and verification (V\&V) techniques applied to their software? Such techniques are widespread in areas such as
aeronautic, railway, etc, but are still seldom used in robotics.  Nowadays, robotic software developments use model-based or model-driven
software engineering approaches (e.g. \textsc{SmartSoft}~\cite{Schlegel:2009tg}, RobotML~\cite{Dhouib:2012wa},
MontiArc~\cite{Ringert:2015us}). These approaches and their associated middleware are numerous and surveyed in a number of
papers~\cite{Brugali:2015eg,Elkady:2012jt,Mohamed:2008ec}. Still, most of theses approaches remain disconnected to formal model analysis and
the use of V\&V techniques. Several works proposed to use the formal synchronous language ESTEREL~\cite{Boussinot:1991wz} to model
functional components ~\cite{Espiau:1996vu,Sowmya:2002ct,Kim:2005gi}. The formal models were then exploited to verify behavioral and timed
properties using model checking tools.  These experiments were nevertheless led on simple examples and specifications were either hard-coded
in ESTEREL or manually translated from robotic components. More recently a special issue~\cite{KressGazit:2011hj} on this subject presented
a number of interesting works along hybrid automata~\cite{Muradore:2011ka} and controller synthesis~\cite{KressGazit:2011bh} and more
recently~\cite{Jing:2016vu}.  The formal frameworks proposed are similar to the ones we use, but they are mostly deployed at decisional
level or on rather simple robotic systems. Our approach is somewhat complementary in choosing to model all the software components which
need to be integrated together at the functional level and then checking and enforcing properties on the integrated code. The closest work
to our approach is MAUVE~\cite{Gobillot:2016hr}, where the code is instrumented to collect Worst Case
Execution Time (WCET), and then temporal formal property can then be checked on the components.  Our approach distinguishes itself by,
simultaneously, being fully automatic, considering all timing constraints of the model and tackling rather complex integrated robotic
applications.

In this paper we present the \GenoMMM{} framework to specify and deploy robotic functional components.  \GenoMMM{} relies on a
specification language which allows the programmer to completely define how the component will work when associated to the user provided
code, and how this code will be internally organized along services, tasks, ports, etc. This specification is done independently of the
middleware targeted and, thanks to its template mechanism, can also be used to synthesize more than just the final code of the component,
but also client libraries, and models for various V\&V frameworks.

The paper first describes, with an example, the specification language used by \GenoMMM{}. Section~\ref{sec:templates} presents the template
mechanism used to synthesize middleware specific components, but also to produce the components 
formal models for two V\&V frameworks. Section~\ref{sec:example} briefly presents two examples (a drone flight controller and an outdoor robot navigation) which
have been completely specified in \GenoMMM{}, followed by a section where we illustrate the type of properties we can formally prove on the
resulting models, as well as how we synthesize a runtime controller to run the component. The paper concludes on the ongoing work along
these lines and the future extensions we intend to tackle.

\section{\GenoMMM{} components}
\label{sec:genom3}

\GenoMMM{}~\cite{Mallet:2010ci} is a tool to specify and implement robotic functional components. In the overall LAAS
architecture~\cite{Ingrand:2007ug}, functional components act as ``servers'' in charge of functionalities which may range from simple
low-level driver control (e.g. the velocity control of the propellers of a drone, camera, etc) to more integrated computations
(e.g. Simultaneous Localization And Mapping (SLAM), navigation, PRM or RRT motion planning, etc).

\subsection{Requirements}
We consider that a typical component is a \emph{program} which needs to handle and manage the following aspects:

\begin{figure}[!tb]
  \centering
  \includegraphics[width=0.9\columnwidth]{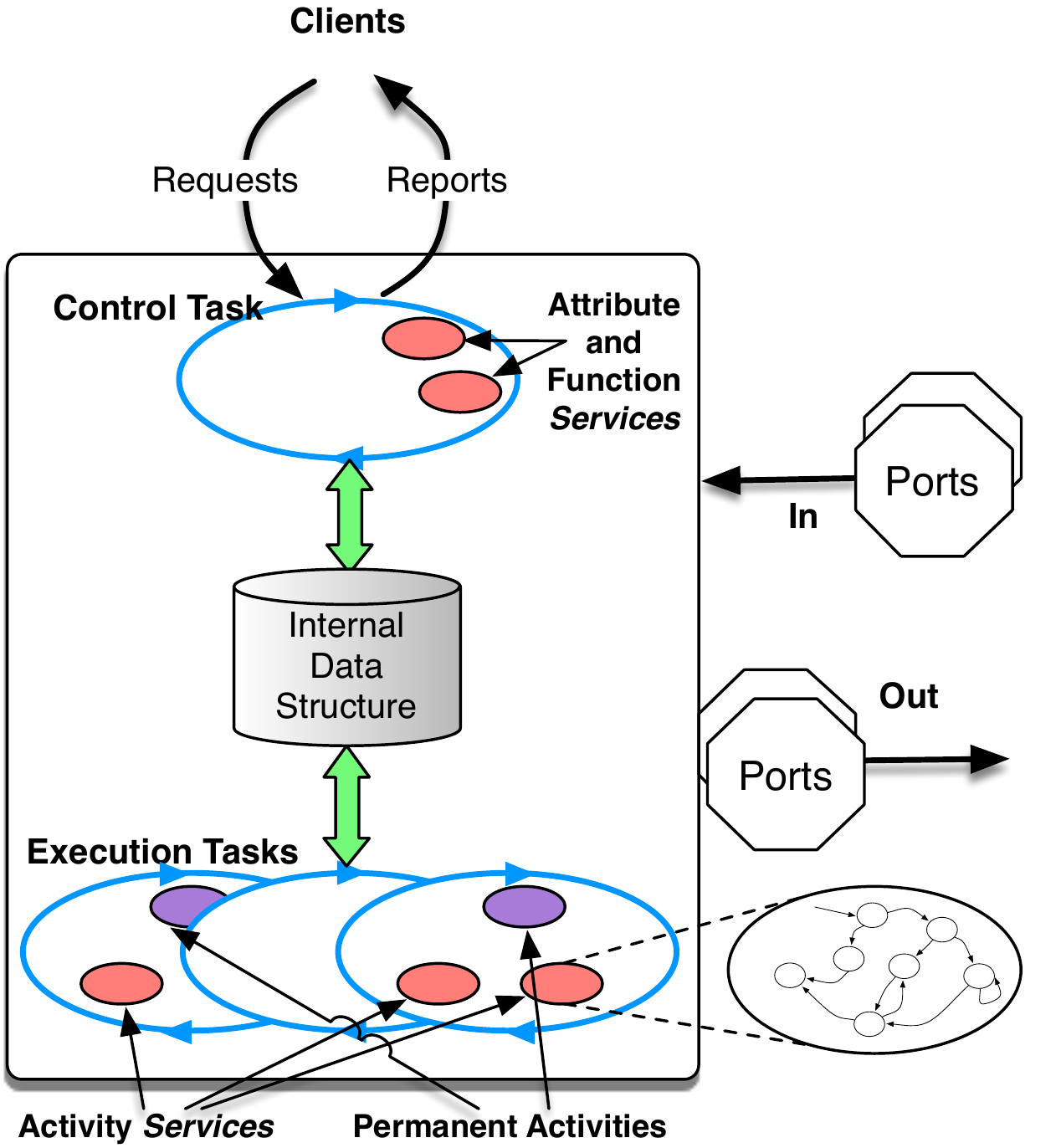}
    \caption{A generic \GenoMMM{} component. \label{fig:gen3}}
\end{figure}

\begin{description}[leftmargin=4mm]
\item[Inputs and Outputs]: a component interacts with external clients and other components. For the former, the control flow, it must handle
  \textit{requests} from client(s) and asynchronously send back \textit{reports}  to the client which issued the request, to act on the result. For the
  latter, the data flow, it must provide a mechanism to share data with other components and read data from other components. Data flow and
  control flow are semantically different and correspond to two different ways components can interact.
\item[Algorithms]: the core algorithms needed to implement the functionality the component is in charge of  must be appropriately organized
  within threads as to preserve the reactivity of the component and the schedulability of the various possibly concurrent algorithms. A component may have
  just one service to provide, but most of the time, there are a number of such services associated to the considered robotic
  functionality. The way algorithms are specified and organized in a component is a tradeoff. One can let the programmer organize its code
  the way it pleases him. But without any particular structure, chances are that little can be validated or verified. If one provides
  guidelines and rules as how the code must be organized, than we stand a much better chance as to prove some properties on the code.
\item[Internally shared data]: the various algorithms, possibly concurrent, running in the component, may have to share state
  variables, parameters, etc. which represent the internal state of the component.
\end{description}

\subsection{Implementation}
\begin{figure}[!ht]
    \centering
    \includegraphics[width=0.75\columnwidth]{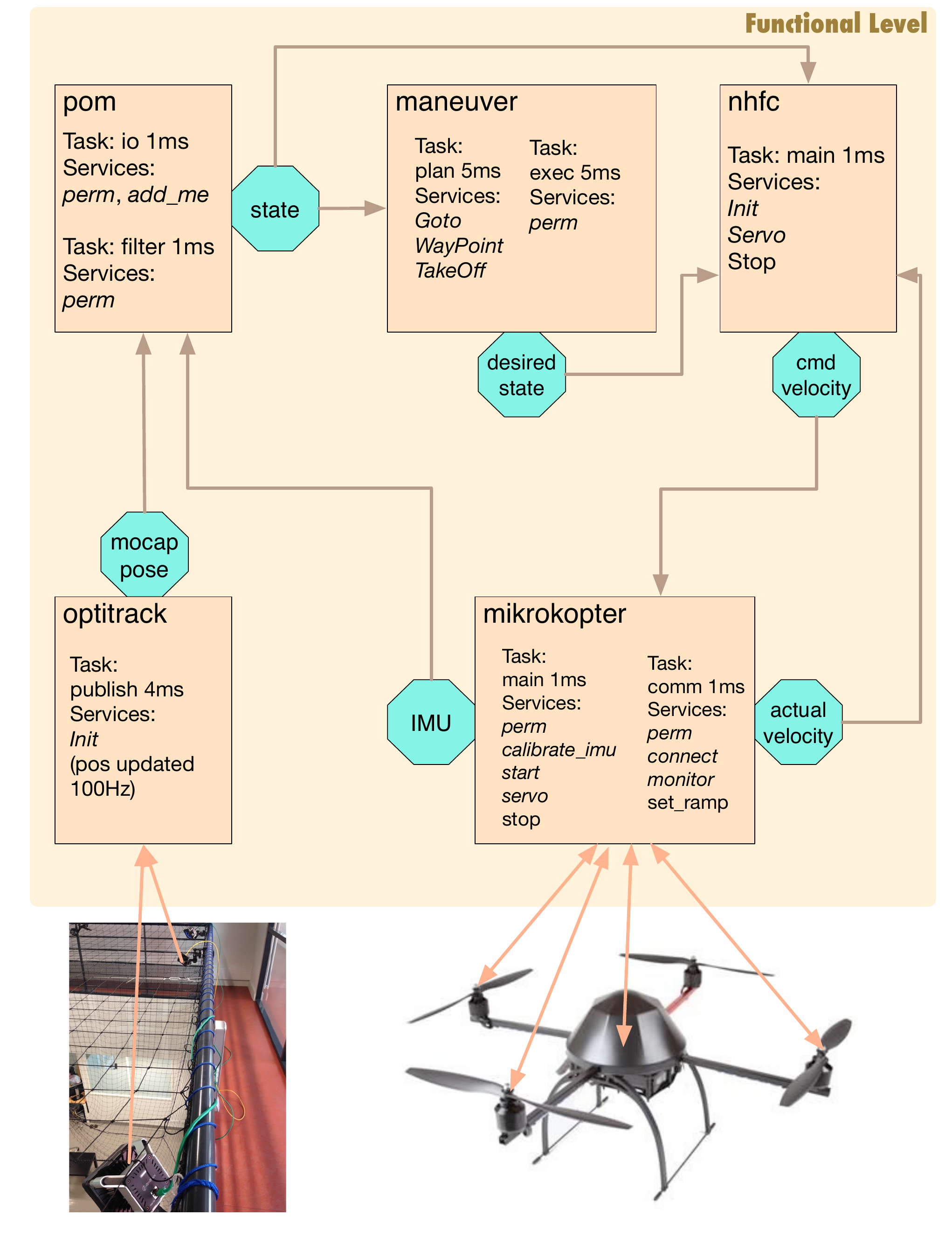}
    \caption{The quadcopter functional level. For the sake of simplicity, only a partial list of services is presented.}
    \label{fig:mkk}
\end{figure}

To achieve such requirements of a functional component, we propose to organize each one along the structure shown on Fig.~\ref{fig:gen3}.  Of course, the
pros and cons of these implementation choices can be discussed, and we will attempt to justify them as we present them. We illustrate the
detailed specifications on Listing~\ref{lst:cpt}, with the \modu{maneuver} component of the complete functional layer of a Mikrokopter drone
flying in our lab (Fig.~\ref{fig:mkk}). This component, given a position to hoover or waypoints to pass by, is in
charge of producing intermediate positions to navigate to.

\begin{lstlisting}[caption={Excerpt from the \GenoMMM{} specification of the \modu{maneuver} component.}, label={lst:cpt},captionpos=b,language=GenoM,showstringspaces=false,columns=fullflexible,numbers=left,xleftmargin=10pt,float=tp]
component maneuver {
  /* ports declaration: direction type name */
  port in	or_pose_estimator::state state;(*@\label{lst:cpt:port}@*)
  port out	or_pose_estimator::state desired;

  /* exception declaration */
  exception e_nostate; /*Service can throw this exception*/(*@\label{lst:cpt:exception}@*)

  /* ids declaration */
  ids { (*@\label{lst:cpt:ids}@*)
    planner_s planner;
    struct trajectory_s { ... } trajectory; ... };

  /* tasks declaration */
  task plan { /* an aperiodic task */(*@\label{lst:cpt:t1}@*)
    codel<start> mv_plan_start(out ::ids) yield ether wcet 26us; /* used to initialize one */ (*@\label{lst:cpt:t1p}@*)
    codel<stop> mv_plan_stop(inout ::ids) yield ether wcet 12us; }; /* used to cleanup once */

  task exec { /* A task with only a permanent activity */ (*@\label{lst:cpt:t2}@*)
    period 5 ms;(*@\label{lst:cpt:per}@*)
    codel<start> mv_exec_start(out desired) yield wait wcet 17us;(*@\label{lst:cpt:perm1}@*)
    codel<wait> mv_exec_wait(in state, in trajectory) yield pause::wait, main 1165us;
    codel<main> mv_exec_main(in state, inout trajectory, out desired) yield wait, pause::main, start wcet 1313us;
    codel<stop> mv_exec_stop() yield ether wcet 4us; };(*@\label{lst:cpt:perm2}@*)

  /* atribute services declaration */
  attribute get_planner(out planner);(*@\label{lst:cpt:as}@*)
  /* function services declaration */
  function stop(out double x, out double y, out double z) {(*@\label{lst:cpt:argout}@*) (*@\label{lst:cpt:fs}@*)
    doc		"Stop a goto and return the current position";
    codel get_current_position(out x, out y, out z) wcet 10us;
    interrupt goto; };

  /* activity services declaration */
  activity goto(in double x, in double y, in double z, in double yaw) {(*@\label{lst:cpt:goto}@*) (*@\label{lst:cpt:argin}@*)
    doc		"Reach a given position from current state";
    task	plan; /* goto will execute in the plan task */ (*@\label{lst:cpt:taskdec}@*)
    validate validate_goto(in z) wcet 3us; /* check z>0 */(*@\label{lst:cpt:validate}@*)
    codel<start> mv_current_state_start(in state, out start) yield plan wcet 43us; (*@\label{lst:cpt:gotoc1}@*)
    async codel<plan> mv_goto_plan(in planner, in start, in x, in y, in z, in yaw, out path) yield exec wcet 2487us;
    codel<exec> mv_plan_exec(in planner, in path, inout trajectory) yield pause::exec, wait wcet 304us;(*@\label{lst:cpt:pause}@*) 
    codel<wait> mv_plan_exec_wait(in trajectory) yield pause::wait, ether wcet 13us;
    codel<stop> mv_plan_exec_stop(out trajectory) yield ether wcet 10us; (*@\label{lst:cpt:gotoc2}@*) (*@\label{lst:cpt:stop}@*) 
    throw e_nostate; /* No valid state position found */(*@\label{lst:cpt:throw}@*)
    interrupt goto; }; /* will interrupt an active goto */(*@\label{lst:cpt:inter}@*)

  activity waypoint(in double x, in double y, in double z, in double yaw) { (*@\label{lst:cpt:wp}@*)
    doc		"Push a given position to reach after last one";
    task	plan; /* waypoint will execute in this task */
    codel<start> mv_waypoint_start(in state, in trajectory, out start) yield plan wcet 6us;
    async codel<plan> mv_goto_plan(in planner, in start,in x, in y, in z, in yaw, out path) yield exec 2487us;
    codel<exec> mv_waypoint_add(in planner, in path, inout trajectory) yield ether wcet 369us; }; };
\end{lstlisting}

Specifying a component in \GenoMMM{} is the programmer design choice. Thus, there are a number of considerations she/he has to take into
account, which may depend on the hardware constraints, the complexity of the algorithm, the needed external data, etc.  Let us describe
these different elements in more details and how they interact, and how they are specified. Note that at this point, we are not assuming any
particular middleware for interprocess communication or data sharing (control and data flow).

Apart from the control task,
each element of the specification results from choices the developper has to carefully make (how many tasks? periodic or not? if periodic, which period? which
services associated to which task? how to break down long services in processing steps? etc):
\begin{description}[leftmargin=4mm]
\item[Control Task]: A component always has a \emph{control task} that manages the control flow by processing \emph{requests} and sending
  \emph{reports} (from/to external clients); activate and stop services, etc. The control task is implicitly comprised within a component and the user 
  does not need to specify it.
\item[Execution Task(s)]: Aside from the \emph{control task}, whose reactivity must remain short, one may need one or more \emph{execution
    tasks}, aperiodic or periodic, in charge of longer computations.
\item[Services]: The core algorithms needed to implement the functionality the component is in charge of are encapsulated within
  \emph{services}. \emph{Services} are
  associated to a \emph{request} (with the same name), but one may also define \emph{permanent activities} which are attached to an \emph{execution task}.
\item[IDS]: A local \emph{internal data structure} is provided for all the \emph{services} to share parameters, computed values or state variables of the
  component. It is appropriately accessed (i.e. with proper locking) by the \emph{services} when they need to read or write a field of the
  \emph{IDS} (line~\ref{lst:cpt:ids}).
\item[Ports]: They specify the shared data, in and out, the component needs or produces from/for other components (line~\ref{lst:cpt:port}).
\item[Exceptions]: One may specify \emph{exceptions}, which can be returned by \emph{services} to \emph{report} non nominal execution
  (line~\ref{lst:cpt:exception}).
\end{description}

We go in more details and see how these different elements interact and how the component internally runs.

\paragraph{Services}\label{services}
Services hold the specifications of the algorithms handled by the component. Services can take arguments (line~\ref{lst:cpt:argin}), and
return values  (line~\ref{lst:cpt:argout}). Services are
activated upon receiving the corresponding request. A service may have a \codel{validate} codel (line~\ref{lst:cpt:validate}). This codel
is executed by the control task and checks that the arguments of the request are correct. If they are, the service is then \textit{runnable},
otherwise, it is reported with an illegal arguments report. A service may also specify other services it interrupts
(line~\ref{lst:cpt:inter}) when it becomes runnable. The interrupted services execute a \codel{stop} codel (line~\ref{lst:cpt:stop}) if any
and report to their client that they have been interrupted.

\emph{Control Services}, are only for short execution as to not delay the control task which executes them. A control service 
may be an \textit{attribute} (setter or getter of fields of the IDS, line~\ref{lst:cpt:as}), or a \textit{function} (line~\ref{lst:cpt:fs}) for quick and simple computations. A \GenoMMM{}
component offers four predefined function services, namely: \serv{Kill} (stop the component), \serv{Abort} (stop an activity service),
\serv{Connect Port} to connect a local \emph{in} port to a distant \emph{out} port and \serv{Connect Service} to connect
a service of another component. 

\begin{figure}[!tb]
    \centering
    \includegraphics[width=0.95\columnwidth]{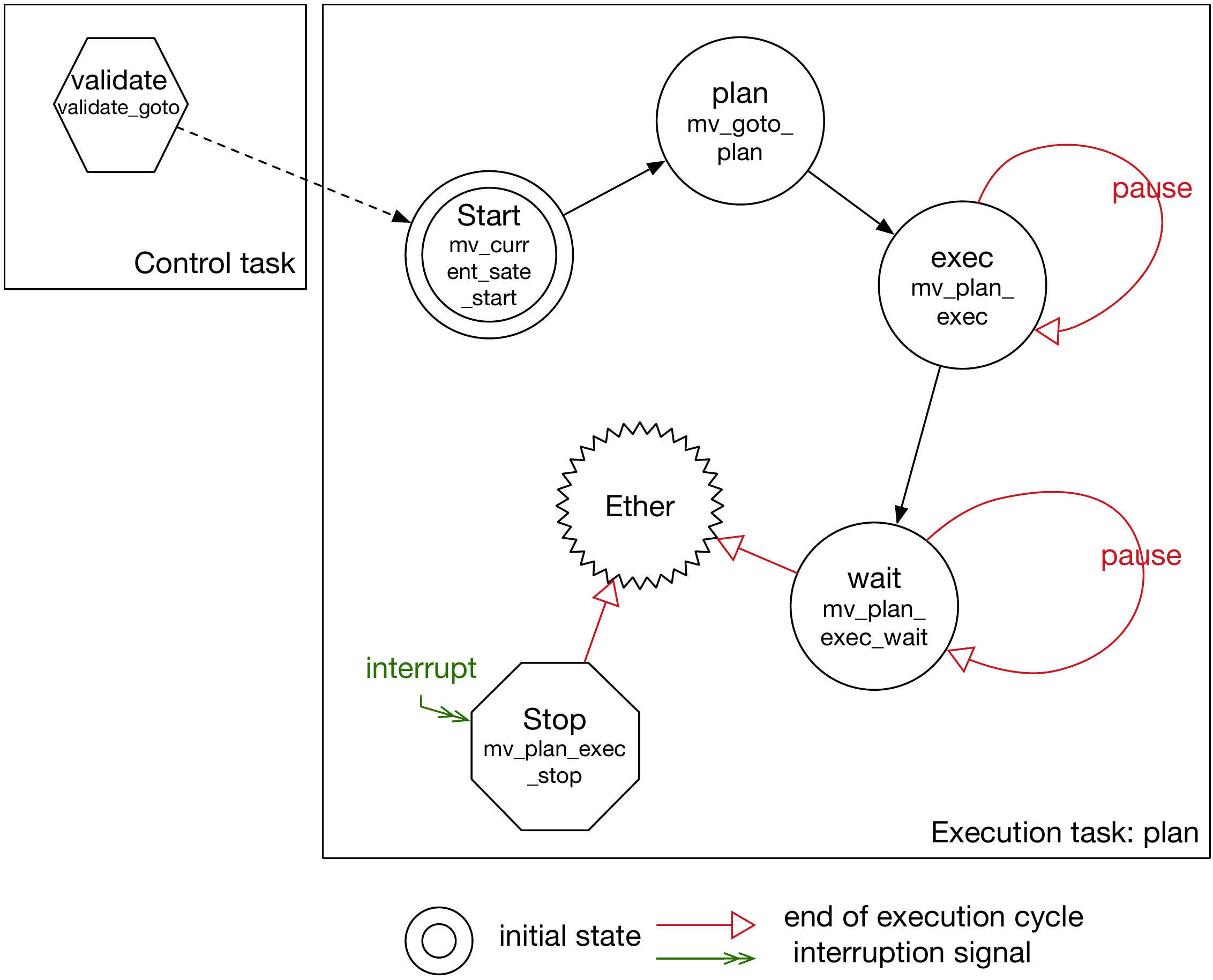}
    \caption{FSM of the \codel{goto} activity of the \modu{maneuver} component (See Listing~\ref{lst:cpt}, line~\ref{lst:cpt:gotoc1}-\ref{lst:cpt:gotoc2})}
    \label{fig:aut}
\end{figure}

\emph{Activity services} (activities for short), see line~\ref{lst:cpt:goto} and~\ref{lst:cpt:wp}, are executed in the execution task
specified in their declaration (line~\ref{lst:cpt:taskdec}). They all have a \codel{start} codel which is the entry point of their codels
finite-state machine (FSM) and as many states/codels, as the programmer wants, to specify the decomposition of the long computation they are performing (e.g., the
FSM defined from line~\ref{lst:cpt:gotoc1} to~\ref{lst:cpt:gotoc2}, also drawn on Fig.~\ref{fig:aut}). The execution of a codel always returns the next state to which the
execution must transition to in the service FSM. If the returned state is prefixed with \emph{pause} (line~\ref{lst:cpt:pause}), the
control of the execution task is passed to the next service to execute in this task, if any, or back to the scheduler as to wait until the
next task period. \emph{ether} is a special state to which a terminating service can transition.  An activity may be \textit{permanent}
(from line~\ref{lst:cpt:perm1} to~\ref{lst:cpt:perm2}). It is not requested by a client and is run by its execution task when the component
starts.

\emph{Codels} specify the C or C++ function they will call, with the arguments (taken from the service arguments, the IDS and/or the ports of the
component) they need (in and out). Codels are restricted to use these arguments only. Codels are also restricted to return a state/codel
specified in the FSM definition of their service.  Each codel may specify a WCET, which measures the worst case execution time of the codel
alone (i.e., executing independently of any other execution). The organisation of activity services along FSM and codels may be seen as an unnecessary
burden on the robot programmer, but nothing prevent the programmer to have one \codel{start} codel which does it all. Yet, breaking code
along a FSM brings a number of advantages when it comes to better code integration and V\&V. It improves
schedulability and code execution interleaving. It provides a finer model of data sharing and code interlocking.

\paragraph{Control Task}
As seen before, the control task manages the requests and reports of the component, as well as starting, terminanting services. It runs the
\codel{validate} codels for services which specify one. If there exist activities that are incompatible with the requested service, the control task instructs the execution tasks in
charge of such activities to interrupt them. 
If the request concerns a control service (attribute, line~\ref{lst:cpt:as} or function, line~\ref{lst:cpt:fs}), the control task executes
it directly. Otherwise, the requested activity service is then put on hold until all the incompatible instances are correctly
interrupted and terminated. Then the control task advises the execution task declared by the service to run it, and sends an intermediate
reply to the client to inform it that processing has started. Upon completion of services, the control task sends reports to the
corresponding client (service ended nominally, service interrupted, etc.).

\paragraph{Execution Tasks}
Execution tasks are periodic (with a specified period, line~\ref{lst:cpt:per}) or aperiodic (line~\ref{lst:cpt:t1}).  With each period signal (if periodic) or event occurrence (if
sporadic), the execution task runs its permanent activity (if any) and then all the \textit{active instances} of its associated activities.
An active instance of a given activity is an instance that has been requested by a client and whose execution has not yet ended.

\paragraph{Internal Data Structure}\label{IDS}
Access to the IDS is mutually exclusive. One can see that the proper specifications (enforced by \GenoMMM{}) of the codel arguments allows
for a very fine grain locking of the IDS field. In other words, we know at any time which codels access what. Only the needed field(s) by a codel 
are locked in order to ensure maximal parallelism. 

\paragraph{Ports}
Information exchange with other components is made through ports (line~\ref{lst:cpt:port}). As seen above, ports usage (in, out or in/out) is also declared in codels
arguments. As such, over a large set of components composing a robotic functional layer, we have a clear model of which codels use a
particular port and at what time.

\section{Templates Approach}\label{sec:templates}

As seen above, \GenoMMM{} provides a rich language to specify functional components and how they should be organized. Still, producing the
real component code to run on the robot, out of this specification, requires additional steps. This is where \GenoMMM{} \emph{template}
mechanism is critical. \GenoMMM{} without template just analyzes the specification file and checks it for
inconsistencies. The real power of \GenoMMM{} is to call it on a specification file, along with a template, as to automatically synthesize
the target of the template. 

\begin{lstlisting}[caption={A  simple template code snippet.}, label={lst:tmp},captionpos=b,language=C,showstringspaces=false,columns=fullflexible,float=!htp]
void
genom_<"$comp">_activity_report(
  struct genom_component_data *self, 
  struct genom_activity *a)
{
  switch(a->sid) {
    case -1: return; /* permanent activity reports nothing */
<'foreach s [$component services] {'>
    case <"$COMP">_<"[$s name]">_RQSTID:
      genom_<"$comp">_<"[$s name]">_activity_report(
        self, 
        (struct genom_<"$comp">_<"[$s name]">_activity *)a);
      return;
<'}'>
  }
\end{lstlisting}

\begin{lstlisting}[caption={Excerpt of the synthesized C code for the PocoLibs \modu{maneuver} component corresponding to the template in Listing~\ref{lst:tmp}
(note how the C code is synthesized for all the services of the component).}, label={lst:code},captionpos=b,language=C,showstringspaces=false,columns=fullflexible,float=!htp]
void
genom_maneuver_activity_report(
  struct genom_component_data *self, 
  struct genom_activity *a)
{
  switch(a->sid) {
    case -1: return; /* permanent activity reports nothing */
    case MANEUVER_connect_port_RQSTID:
      genom_maneuver_connect_port_activity_report(
        self, 
        (struct genom_maneuver_connect_port_activity *)a);
      return;
...
    case MANEUVER_goto_RQSTID:
      genom_maneuver_goto_activity_report(
        self, 
        (struct genom_maneuver_goto_activity *)a);
      return;
    case MANEUVER_waypoint_RQSTID:
      genom_maneuver_waypoint_activity_report(
        self, 
        (struct genom_maneuver_waypoint_activity *)a);
      return;
  }
...
\end{lstlisting}

A template when called by \GenoMMM{} on a given component specification has access to all the information contained in the specification file
such as services names and types, ports and IDS fields needed by each codel, execution tasks periods, etc. Based on all these specifications,
\GenoMMM{} can also compute information such as which codels can execute at the same time (considering their respective arguments), or which
port must be locked by which codel. etc. 
Through the template interpreter
(Tcl), one specifies what they need the template to synthesize. Since the interpreter relies on a complete scripting language, there is
virtually no restriction on what a template can generate. For instance, Listing~\ref{lst:tmp} shows an excerpt of a template function and
Listing~\ref{lst:code} the C code it produces when called together with the \modu{maneuver} specification.  The interpreter evaluates
anything enclosed in \textit{markers} $<$\textsf{'\hspace{1em}'}$>$ without output, while on the code between $<$\textsf{"\hspace{1em}"}$>$,
variables and commands substitution is performed and the result is output in the destination file, together with the text outside of the markers. For example,
\verb|<'foreach s [$component services] {'>... <'}'>|
iterates over the list of services of the component, contained in the \verb|$component| variable; while \verb|<"[$s name]">| is replaced by
the name of the service contained in the \verb|$s| variable bound by the foreach statement.

There are a number of templates already defined, to synthesize the component code for a given middleware (e.g., PocoLibs\footnote{\url{https://git.openrobots.org/projects/pocolibs}}, ROS~\cite{Quigley:2009tg}), C client
libraries, OpenPRS client procedures and code,  a JSON client, etc. For example, the template \textit{skeleton} generates the files containing the
codel stubs with their proper function prototypes. The user can then specify the algorithmic core of their codels without worrying about the
middleware.

\subsection{Middleware Independence}

The middleware ``server'' templates are used to synthesize the component itself to be run on the robot. They output the \textit{glue code}
in charge of making calls to the targeted middleware. The synthesized glue code manages message passing (requests/reports) as well as ports
connection, and handles all the internal algorithms to manage the different tasks, services FSMs, proper locking of shared resources,
etc. This is a viable solution to the problem of middleware dependency as neither the specification nor the codels rely on a specific
middleware. Indeed, codel execution can only rely on the objects declared in their arg list (i.e. IDS fields, their service arguments and ports) and do not make
calls to the middleware. 

So far, \GenoMMM{} offers several middleware templates, notably ROS-Com and
PocoLibs. The former heavily uses ROS topics, ROS services and
ROS actions (actionLib) in the synthesized code, while the latter uses PocoLibs primitives, such as MBox, CSMBox, posters, etc. But from an
external behavior point of view, the two resulting components (ROS and PocoLibs) behave exactly the same (apart from performance issues
specific to the middleware implementation).

\subsection{Formal Verification}

\GenoM{} started as a robotic software development tool and methodology in the mid 90's~\cite{Fleury:1997ta}. Quickly, it appeared that the
component specification could be used for more than components code synthesis. An earlier study, using \GenoM{}2, did an ad hoc job at
synthesizing a BIP~\cite{Basu:2006ve} model and went as far as running the synthesized model of 14 components along with the BIP Engine
on a real robot~\cite{Bensalem:2011uf}. \GenoMMM{} has a semantically cleaner specification model (arbitrarily complex user-defined
FSMs, codels WCETs, etc) and thanks to its versatile template mechanism can now be used to synthesize formal models for different frameworks (\fcr~\cite{Berthomieu:2008vo}
and  Real-Time BIP (RT-BIP)~\cite{Abdellatif:2010tv}).  But one should keep in mind that we want to synthesize the model which is semantically equivalent to the resulting
component. So the synthesized model goes beyond what is in the component specification file. As a
result, the Fiacre or RT-BIP templates synthesize models relative to a targeted specific middleware. Moreover, unlike the previous work
presented in~\cite{Bensalem:2011uf}, the current \GenoMMM{} component specifications include some temporal information: WCET on codels, as
well as the period of execution tasks. WCET can be obtained empirically, or with more advanced techniques~\cite{Wilhelm:2008gu}, for
now, we get them by running the components and collecting data. Yet, these temporal information were not taken into account in~\cite{Bensalem:2011uf} but are now part
of the formal models we synthesize.

In~\cite{Foughali:2016tma}, a template is presented which automatically synthesizes models in \fcr~\cite{Berthomieu:2008vo}, a formal language
for specifying concurrent and real-time systems based on automata (behavior) and time Petri nets (timing aspects). The synthesized models
are exploited in order to verify important real-time properties using \tina~\cite{Berthomieu:2004dp} model checkers.

\GenoMMM{} also provides a template to automatically generate RT-BIP~\cite{Basu:2006ve} models. RT-BIP is a formal framework
based on interacting components encapsulating timed automata with urgencies. It is associated to an execution \textit{engine} and an offline deductive verification tool:
RTD-finder~\cite{BenRayana:2016wd}.  The latest release of the RT-BIP engine (RT-BIPE) implements \textit{external}
transitions guarded with external, non controllable events.  This allowed us to run the generated models with the RT-BIPE checking for
client requests while
properly handling the execution of sporadic tasks, particularly the control task, and monitor our components online.

\section{Deployed Examples}
\label{sec:example}

There are already a number of experiments deployed with \GenoMMM{}. We illustrate our approach with the functional level of \textit{(i)} a MikroKopter
quadcopter flying in our lab and \textit{(ii)} our outdoor robot Mana. 

\subsection{Quadcopter Flying Example}
\label{sec:example:quad} 
Fig.~\ref{fig:mkk} presents the 5 components involved in our quadcopter functional layer. Each box corresponds to a component, and
each octagon is a port. Ports are written (out) by the components they are attached to, and read (in) through the arrow pointing to
the reading component. Inside each box, we list the execution tasks (their period or ``ap'' if they are aperiodic), and a partial list of
the services provided by this component. Note that this figure does not present the ``supervisor'' in charge of sending requests and
analyzing reports, which is out of scope of this paper.
\begin{itemize}[leftmargin=5mm]
\item \modu{mikrokopter} is the component in charge of the quadcopter low-level hardware. The quadcopter is controlled by applying a
  velocity to each propeller, and produces the current velocities, as well as its current IMU (Inertia Measurement Unit) values. It has two
  tasks i) \task{comm}, aperiodic, which keeps polling and parsing data from the hardware (to get the current propellers velocity and IMU) and
  storing them in the IDS. ii) \task{main}, periodic at \unit{1}{ms}, which reads the \port{cmd velocity} port, manages the servo control and
  writes the two ports \port{IMU} and the propellers \port{actual velocity}.
\item \modu{optitrack} is the component handling the current position of the quadcopter as perceived by our ``OptiTrack'' motion capture
  system. It has one task \task{publish}, periodic at \unit{250}{Hz}. It provides the
  current position of the quadcopter in the \port{mocap pose} port.
\item \modu{pom} merges the \port{mocap pose} position produced by \modu{optitrack} and the \port{IMU} from \modu{mikrokopter} and produces an
  Unscented Kalman filtered position in port \port{state}. It has two tasks \task{io} and \task{filter} both periodic at
  \unit{1}{KHz}.
\item \modu{maneuver} is the navigation component, it has two tasks \task{exec} with a period of \unit{5}{ms} and \task{plan} aperiodic. Given a
  position or waypoints to navigate to, it reads the \port{state}, and computes a trajectory to reach it, producing intermediate
  positions to fly to in \port{desired state}.
\item \modu{nhfc} (Near Hovering Flight Controller) is the core of the flight controller. Running one task \task{main} at \unit{1}{KHz}, it
  reads the \port{actual velocity} port of the propellers, the current position in the \port{state} port of \modu{pom}, and
  the desired position (port \port{desired state}) of \modu{maneuver} and produces the proper \port{cmd velocity} port
  containing the desired velocity of the propellers (which is then read by \modu{mikrokopter}) to reach and hover near this position.
\end{itemize}

The complexity of our quadcopter functional level is such that with 5 components, 13 tasks running potentially in parallel, over 40 services
and more than 65 codels, checking by hand any temporal property is impossible. Note that for the sake of simplicity, only a partial list of
services is presented in this paper, we refer the reader to the source repository for the complete specifications
\url{https://git.openrobots.org/projects/telekyb3} (in the corresponding sub-projects).

\subsection{Outdoor Navigation Example}
\label{sec:example:mana}
\begin{figure}[!t]
    \centering
    \includegraphics[width=0.75\columnwidth]{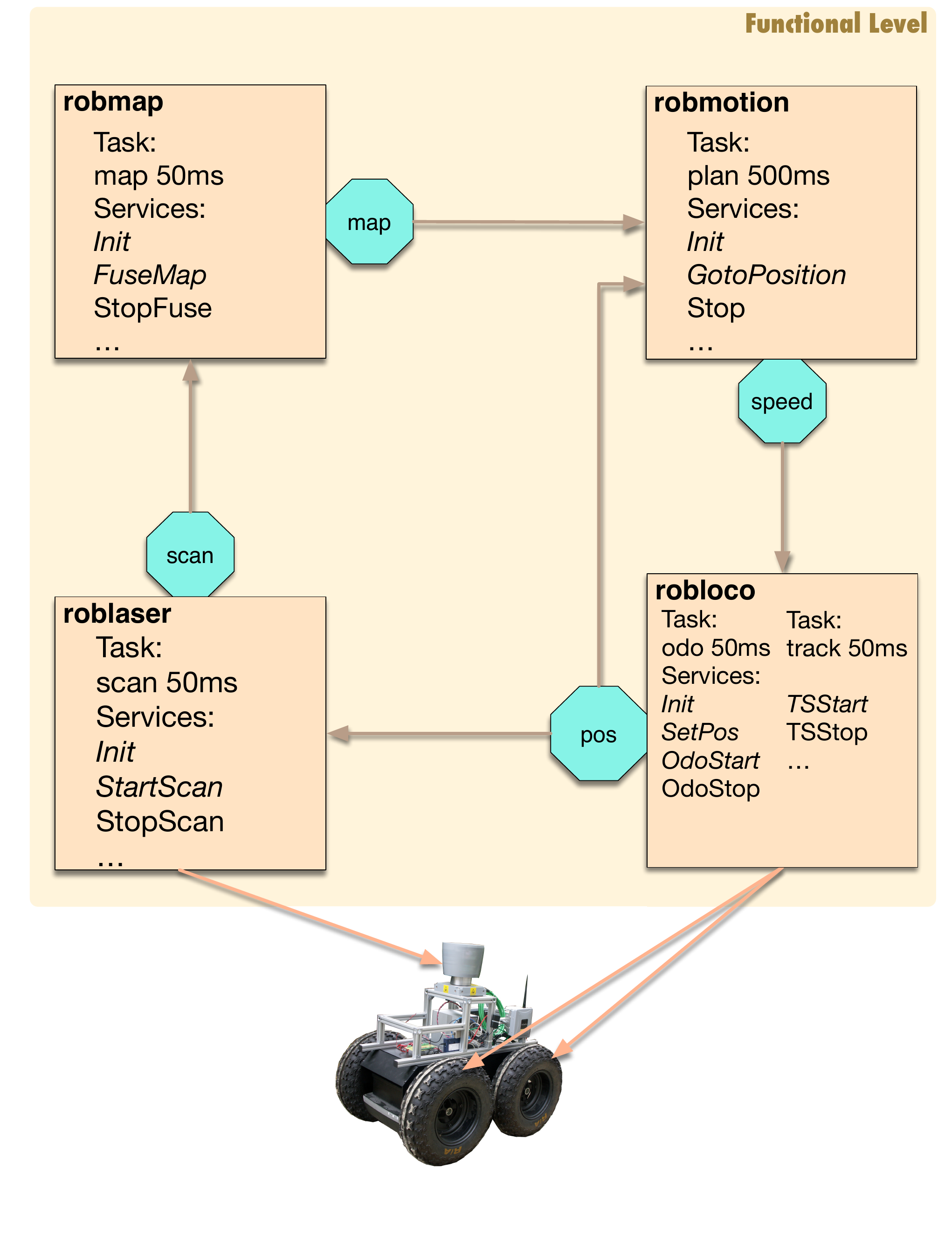}
    \caption{An outdoor robot functional level}
    \label{fig:flm}
\end{figure}

This navigation stack is inspired from the real navigation running on our Segway RMP 400 robot, Mana.

Fig.~\ref{fig:flm} presents the four components in charge of the navigation:

\begin{itemize}[leftmargin=5mm]
\item \modu{robloco} is in charge of the robot low-level controller. It has a \task{track} task (period 50 ms) associated to the
  activity \serv{TSStart} (TrackSpeedStart, interruptible by the function \serv{TSStop}) that reads data from the \port{speed} port and sends
   it to the motor controller. In parallel, one of the \task{odo} task (period 50 ms) associated activities, namely \serv{OdoStart} (interruptible
   by the function \serv{OdoStop}), reads the encoders on the wheels and produces a current position on the \port{pos} port.
\item \modu{roblaser} is in charge of the laser. It has a \task{scan} task (period 50 ms) which runs the \serv{StartScan} activity 
(interruptible by the function \serv{StopScan}). \serv{StartScan} produces, on the port \port{scan}, the free space in the laser's range tagged with the position 
where the scan has been made (read on \port{pos}).
\item \modu{robmap} aggregates the successive \port{scan} data in the \port{map} port. A \task{fuse} task (period 50 ms) and \serv{FuseMap}, one of its activities,
 perform the computation. The function \serv{FuseStop} interrupts the activity \serv{FuseMap}.
\item \modu{robmotion} has one task \task{plan} (period 500 ms) which, given a goal position (via the activity \serv{GotoPosition}), computes the appropriate speed 
to reach it and writes it on \port{speed}, using the current position (from \port{pos}), and avoiding obstacles (from \port{map}). \serv{GotoPosition}
interrupts itself, so a new request will cancel the currently running one (if it exists) and force the execution of its \codel{stop} state. Similarly, The
  \serv{Stop} service (function) interrupts \serv{GotoPosition}.
\end{itemize}

\section{Offline and Runtime Verification}

Templates are used to generate our quadcopter and outdoor robot components for various middleware. We use these components today routinely on PocoLibs and
ROS-Com middleware. In this section, we focus on verification of the PocoLibs implementation as it is more adapted for real-time
applications such as flight control running at \unit{1}{KHz}.

\subsection {Offline Real-time Properties with \fcr}

When it comes to prove real-time properties on concurrent real-time systems,
one has to take into account the hardware on which it runs, and the scheduling policy used. The latest release of the \fcr template allows the user to provide the number of cores
provided by the platform. The template consequently generates a \fcr model of the components including a cooperative (non-preemptive) FIFO scheduler in charge 
of allocating the cores to the different tasks present in the components model\footnote{In other words, the tasks are not preempted, and
  release their core/CPU at the end of each cycle.}. We give hereafter two examples of real-time properties that we successfully 
prove on our, respectively, quadcopter and outdoor components. These properties are of a capital importance to robotic programmers. Note that we automatically synthesize 
the \fcr models of the components and the external clients considering their behavior as they produce the requests as needed for both applications.

\paragraph{Schedulability (Quadcopter)}

We refer to a periodic execution task as \emph{schedulable} if it always executes the requested services before its deadline (next period
signal).  Schedulability is very often a hard real-time requirement in quadcopter applications.
The problem of verifying the schedulability of tasks is inherently complex in \GenoMMM{} because of the mutual exclusion between
codels. For instance, in \modu{mikrokopter}, while the task \task{main} is executing a given service, it may have to wait when it reaches
one of its codels since such a codel needs a resource (a field of the IDS or a port) already in use by a codel being executed by the control
task or the task \task{comm} or even a task of another component since ports are shared among components. Hence, verifying the
schedulability is more complicated than just summing the WCETs of the codels and comparing the result to the task period. Furthermore, the
lack of cores may cause a delay between the period signal and the actual start of task execution, raising the risk to miss the deadline. To
verify schedulability properties, we generate an external client to ensure a stationary flight (hovering).

The modeling choices in~\cite{Foughali:2016tma} permit an easy expression of schedulability properties for all tasks e.g. for \task{main}:

\begin{lstlisting}[language=FIACRE,label={lst:sched},captionpos=b,columns=fullflexible]
property schedulability_main is 
always (microkopter/main/state executing => not (main_period_signal))
\end{lstlisting}
Which translates to: for all execution paths, when \task{main} is at its state \texttt{executing}, the boolean \texttt{main\_period\_signal}
(which is set to true in the model when \task{main} period is reached) evaluates to false. The same modeling choices ease also the
verification of schedulability properties.  Indeed, they are expressed as invariants which allows the use of the \tina coarser reachability
graph construction that does not preserve firing sequences (smaller state spaces).  The results, obtained in about seven minutes, show that all
tasks are schedulable, considering the quad-core hardware constraints. We note that in case of using a less powerful hardware, e.g. a core-duo
platform, the model checker produces a counterexample as the schedulability property holds no more for \task{filter}
(\modu{pom}). Considering the aforementioned scheduler characteristics, we prove that in order to satisfy schedulability properties for the hovering 
application, a hardware with three cores minimum is indispensable. 

\paragraph{Bounded Stop (Outdoor Robot)} \label{safestop}

As soon as a \serv{Stop} request is sent to \modu{robmotion}, the\codel{stop} codel of \modu{robmotion}'s \serv{GotoPosition} is executed
(because the specification of the \serv{Stop} service is to interrupt the \serv{GotoPosition} service). This execution includes writing a
null speed to \port{speed} port that will be sent to the motor controller via executing the codel \codel{update} of \serv{TSStart}
(\modu{robloco}). For the robot programmer, it is naturally important to determine the maximum amount of time $\tau$ between sending a
\serv{Stop} request and applying a null speed to the wheels. Due to mutual exclusion among cores and ports, manually calculating $\tau$
would be as tedious as error-prone.

We make use of the \fcr patterns~\cite{Abid:2012dk} \texttt{leadsto within} and \texttt{leave} to 
compute the worst-case value of $\tau$ considering the actual quad-core platform on Mana. 
 \begin{lstlisting}[language=FIACRE,label={lst:rmp2},captionpos=b,columns=fullflexible]
property bounded_stop_1 is (robmotion/control_task/state Stop_req) leadsto (robmotion/GotoPosition/state stop) within [0,0.5]
property bounded_stop_2 is (robmotion/GotoPosition/state stop) leads to leave (robloco/TSStart/state update) within [0,0.06]
\end{lstlisting}
The pattern \texttt{leadsto} encodes the Linear Temporal Logic (LTL) combined operator $\square$~$\lozenge$ (always eventually). The \textit {scope modifier} \texttt{within} extends the pattern 
 \texttt{leadsto} in order to express a timed property. The pattern \texttt{leave} expresses leaving a given state. Indeed, it is important in this context to compute the bound 
up to the end of \codel{update} execution to make sure the null speed is sent to the wheels controller. We sum the bounds separating (1) sending the \serv{Stop} request 
and reaching the codel \codel{stop} of \serv{GotoPosition} (\modu{robmotion}) and (2) reaching the codel \codel{stop} of \serv{GotoPosition} and leaving the codel \codel{update} 
of \serv{TSStart} (\modu{robloco}) to prove the sought bound $\tau$ to be \unit[560]{ms}.

\subsection{Online Real-time Controller with RT-BIP}

RT-BIP models are automatically generated out of \GenoMMM{} specifications. The \GenoMMM{}-to-RT-BIP template synthesizes readily
executable models (linked with the codels) for the multi-threaded RT-BIPE. This latest release of RT-BIPE, allows the definition of
asynchronous external transitions, which permits us to run the component on the RT-BIPE while communicating directly with external clients. We use
these models to conduct runtime monitoring on the outdoor robot navigation components \modu{robmotion} and \modu{robloco} while evolving with
the other components implemented in PocoLibs.

For instance, the RT-BIPE enforces the timed constraints of the model (respect of task period and WCET) and is able to catch when those
constraints are violated, and take appropriate measures. The multithreaded RT-BIP engine is still currently under development toward a
version that will allow us to run all the components together on the engine and perform runtime verification and online enforcement of
properties.  As for offline verification of the model with RTD-Finder, it is still being investigated as the tool does not yet readily
support urgencies and variables.  RTD-Finder, being a deductive verification tool on invariants automatically extracted from the model
(behavior invariants, interaction invariants and history clock invariants), may be a legitimate resort when models do not scale with model
checking. The outcome of the ongoing work on integrating urgencies and/or variables in the verification process will certainly open a new
horizon in verifying more complex systems in the future.

\section{Conclusion}\label{sec:conclusion}

\GenoMMM{} defines a functional component specification language and offers a template mechanism which allows the programmer to synthesize
the component code ready to be built, linked to the user written algorithms, and to deploy it on the targeted platform. There exist a number
of templates, for example to synthesize the components for various middleware. Thus the middleware choice can be done later with respect to
performance issues, development platform or other independently deployed components and legacy software. Similarly, one can now
automatically synthesize the formal models, for two frameworks widely used by the V\&V community for embedded and concurrent real-time
systems (\fcr and RT-BIP). These formal models can then be used with their respective toolboxes, to formally check offline properties, using
complementary approaches: model checking for \fcr; and invariants extraction and satisfiability for RT-BIP.  RT-BIP, together with its
Engine can also run the formal models linked to the user code, in place of the regular component, and enforce real-time properties at
runtime. From the roboticist point of view, being able to formally check functional components alone, but also integrated with other
components, with two different V\&V paradigms is a win-win situation, as one can choose the most adapted technique to the particular
behavioral and timed properties they want to prove. Moreover, with the rising complexity of autonomous systems (autonomous cars, drones,
cobots, etc), formal analysis and proof of correct behavior is becoming a rather critical issue and can lead to new ways of certifying the
software of these systems. For example, we are now starting a new project to apply this methodology to an autonomous driverless bus, and
expect this approach to ease the certification of such a vehicle, with respect to regulation agency.

As for future work, we plan to take advantage of the upcoming versions of the RT-BIP Engine to be able to run all the components of a functional
layer. We also plan to improve how to have roboticist (not V\&V expert) to express the properties they want to prove and how to easily interpret the results.

\bibliographystyle{IEEEtran}

\end{document}